\documentclass[10pt,twocolumn,letterpaper]{article}

\usepackage{cvpr}
\usepackage{times}
\usepackage{epsfig}
\usepackage{graphicx}
\usepackage{amsmath}
\usepackage{amssymb}
\usepackage{multirow}
\usepackage{stfloats}
\usepackage{float}
\usepackage{authblk}
\usepackage{subfig}
\newcommand{\head}[1]{{\noindent\textbf{#1}}}

\usepackage[pagebackref=true,breaklinks=true,letterpaper=true,colorlinks,bookmarks=false]{hyperref}

\cvprfinalcopy 

\def\cvprPaperID{1278} 

\begin{document}

\makeatletter
\renewcommand\AB@affilsepx{\quad \protect\Affilfont}
\makeatother
\title{Generic Event Boundary Detection: A Benchmark for Event Segmentation}
\author[1,2]{Mike Zheng Shou}
\author[2]{Stan Weixian Lei}
\author[1]{Weiyao Wang}
\author[1]{Deepti Ghadiyaram}
\author[1]{Matt Feiszli}
\affil[1]{Facebook AI}
\affil[2]{National University of Singapore}

\renewcommand\Authands{, }

\maketitle

\begin{abstract}
This paper presents a novel task together with a new benchmark for detecting generic, taxonomy-free event boundaries that segment a whole video into chunks.
Conventional work in temporal video segmentation and action detection focuses on localizing pre-defined action categories and thus does not scale to generic videos.
Cognitive Science has known since last century that humans consistently segment videos into meaningful temporal chunks.
This segmentation happens naturally, without pre-defined event categories and without being explicitly asked to do so.
Here, we repeat these cognitive experiments on mainstream CV datasets; with our novel annotation guideline which addresses the complexities of taxonomy-free event boundary annotation, we introduce the task of \textbf{Generic Event Boundary Detection (GEBD)} and the new benchmark \textbf{Kinetics-GEBD}.
We view GEBD as an important stepping stone towards understanding the video as a whole, and believe it has been previously neglected due to a lack of proper task definition and annotations.
Through experiment and human study we demonstrate the value of the annotations.
Further, we benchmark supervised and un-supervised GEBD approaches on the TAPOS dataset and our Kinetics-GEBD.
We release our annotations and baseline codes at \textbf{CVPR'21 LOVEU Challenge}: \url{https://sites.google.com/view/loveucvpr21}.
\end{abstract}

\section{Introduction}

\begin{figure}[t]
\begin{center}
\includegraphics[width=\linewidth]{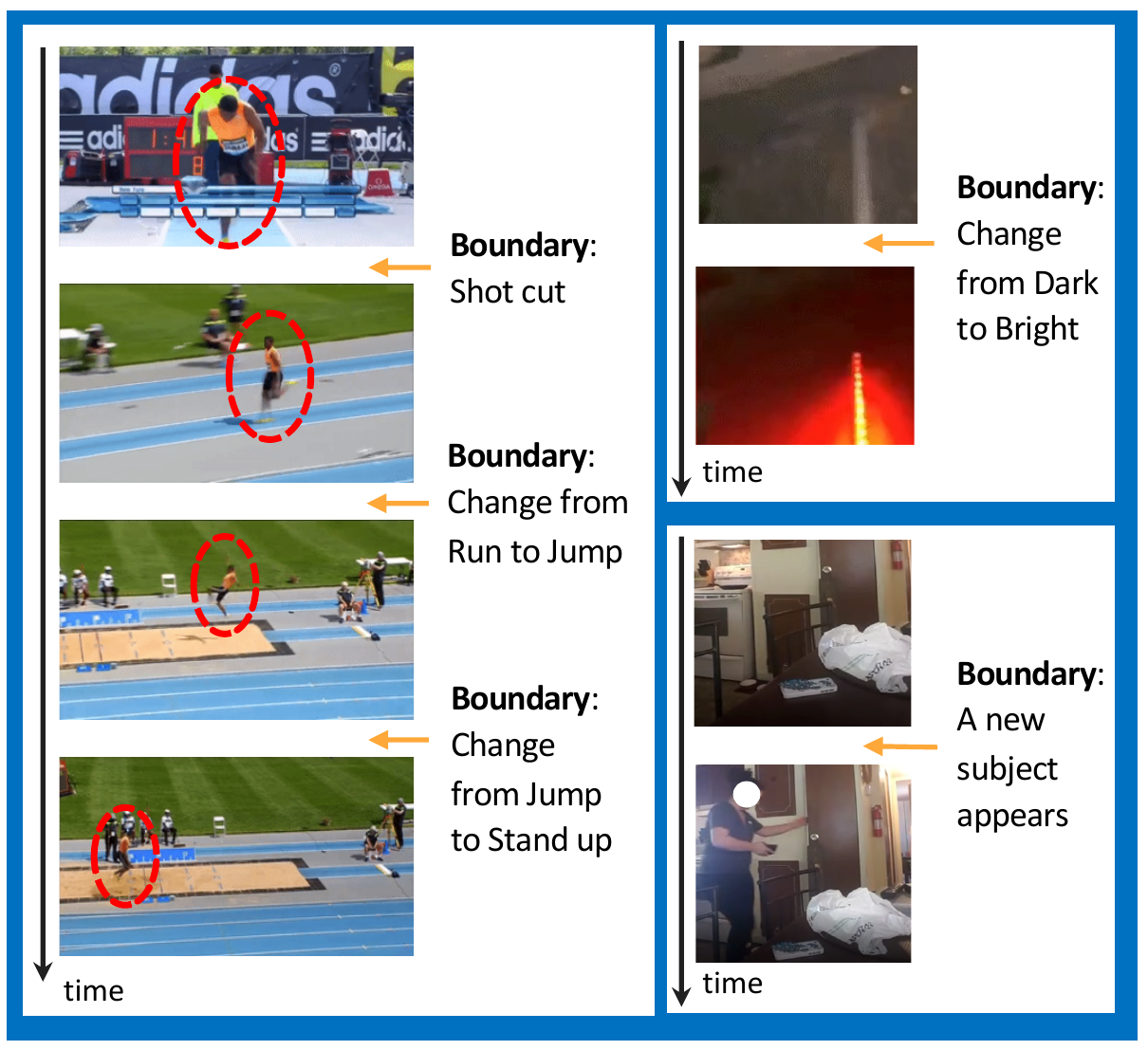}
\end{center}
\vspace{-1em}
\caption{Examples of generic event boundaries: 1) A long jump is segmented at a shot cut, then between actions of Run, Jump and Stand up (dominant subject in red circle).
2) color / brightness change
3) new subject appears.}
\label{fig:teaser}
\end{figure}

\begin{table*}[t]\footnotesize
\begin{center}
\begin{tabular}{c|rrrccrc}
                                       & \#videos & \#segments & \#boundaries & video domain & boundary cause                                               & \#boundary classes  & \#Annotations per video \\ \hline
THUMOS'14                              & 2700     & 18K      & 36K        & sports       & action                                                & 20         & 1                                      \\
ActivityNet v1.3                       & 27801    & 23K      & 46K        & in-the-wild  & action                                                & 203        & 1                                              \\
Charades                               & 67000    & 10K      & 20K        & household    & action                                                & 157        & 1                         \\
HACS Segments                          & 50000    & 139K     & 278K       & in-the-wild  & action                                                 & 200        & 1                         \\
AVA                                    & 214      & 197K     & 394K       & movie        & action                                                 & 80         & 1                         \\
EPIC-Kitchens     & 432      & 39K      & 79K        & kitchen      & action                                                 & 2747, open-vocab       & 1                         \\
EPIC-Kitchens-100 & 700      & 89K      & 179K       & kitchen      & action                                                 & 4053, open-vocab       & 1                         \\
TAPOS Instances   & 16294    & 48K           &     33K         & sports       & action                                                 & open-vocab         & 1                         \\
Kinetics-GEBD (raw)                   & 55351     &     1771K        &     1498K         & in-the-wild  & generic & taxonomy-free & 4.93  \\
Kinetics-GEBD (clean)                   & 54691     &     1561K	        &     1290K         & in-the-wild  & generic & taxonomy-free & 4.94                                          
\end{tabular}
\end{center}
\caption{Comparing our Kinetics-GEBD with other video boundary datasets. Our Kinetics-GEBD has the largest number of temporal boundaries (e.g. 32x of ActivityNet, 8x of EPIC-Kitchens-100), spans a broad spectrum of video domains in the wild in contrast to sports or kitchen centric, is open-vocabulary rather than building on a pre-defined taxonomy, contains boundaries caused by not only action change but also generic event change, and has almost 5 annotations per video to capture human perception differences and therefore ensure diversity. 
Note that for ActivityNet and TAPOS, since ground truths of test set are withheld, we do not include \#segments and \#boundaries of their test sets.
}
\label{table:dataset-compare}
\end{table*}

Cognitive science tells us \cite{tversky2013event} that humans perceive video in terms of ``events'' (goal-directed sequences of actions, like ``washing a car'' or ``cooking a meal''), and further, people segment events naturally and spontaneously while perceiving video, breaking down longer events into a series of shorter temporal units. However, mainstream SOTA video models \cite{tran2015learning,tran2018closer,slowfast,Kinetics,tsm,x3d} still commonly process short clips (e.g. 1s long), followed by some kind of pooling operation to generate video-level predictions.


Recent years have seen significant progress in temporal action detection \cite{chao2018rethinking,dai2017temporal,gao2017cascaded}, segmentation \cite{ST-CNN,alayrac2017joint,kuehne2014language,farha2019ms} and parsing \cite{pirsiavash2014parsing,shao2020tapos} in videos.  Despite this, we have not seen major developments in modeling long-form video.  The cognitive science suggests that one underlying deficit is event segmentation: unlike our SOTA models, humans naturally divide video into meaningful units and can reason about these units.  In contrast to our current methods build upon limited sets of predefined action classes, humans perceive a broad and diverse set of segment boundaries \textbf{without any predefined target classes}.

To enable machines to develop such ability, we propose a new task called \textbf{Generic Event Boundary Detection (GEBD)} which aims at localizing the moments where humans naturally perceive event boundaries.
As Fig.~\ref{fig:teaser} shows, our event boundaries could happen at the moments where the action changes (e.g. Run to Jump), the subject changes (e.g. a new person appears), the environment changes (e.g. suddenly become bright), for example.

To annotate ground truths of such taxonomy-free event boundaries, the common strategies used by the existing temporal tasks with pre-defined taxonomy do not work:
\begin{enumerate}
    \item Existing tasks require us to manually define each target class carefully i.e. its semantic differentiators compared to other classes. But it is impractical to enumerate and manually define all candidate generic event boundary classes.
    \item The existing tasks typically focus on shot and action boundaries, neglecting other generic event boundaries as the examples shown in Fig.~\ref{fig:teaser} like change of subject.
\end{enumerate}
In this paper, we propose to follow cognitive experiments \cite{tversky2013event} in annotating event boundaries on computer vision datasets.
We choose the popular Kinetics \cite{kay2017kinetics} dataset as our video source and construct a new event segmentation benchmark \textbf{Kinetics-GEBD}.
The marked boundaries are relatively consistent across different annotators; the main challenge raising ambiguity is the level of detail. For example, one annotator might mark boundaries at the beginning and end of a dance sequence, while another might annotate every dance move.
We develop several novel principles in design annotation guideline to ensure consistent level of detail across different annotators while explicitly capturing the human perception differences with a multi-review protocol.


Our new GEBD task and benchmark will be valuable in:
\begin{enumerate}
    \item Immediately supporting applications like video editing, summarization, keyframe selection, highlight detection.  Event boundaries divide a video into natural, meaningful units and can rule out unnatural cuts in the middle of a unit, for example.
    \item Spurring progress in long-form video; GEBD is a first step towards segmenting video into meaningful units and enabling further reasoning based on these units.
\end{enumerate}
.





In summary, our contributions are four-fold:
\begin{itemize}
\item A new task and benchmark, Kinetics-GEBD, for detecting generic event boundaries without the need of a predefined target event taxonomy.
\item We propose novel annotation task design principles that are effective yet easy for annotators to follow.  We disambiguate what shall be annotated as event boundaries while preserving diversity across individuals in the annotation.
\item We benchmark a number of supervised and un-supervised methods on the TAPOS \cite{shao2020tapos} dataset and our Kinetics-GEBD. 
\item We demonstrate the value of our event boundaries on downstream applications including video-level classification and video summarization.
\end{itemize}

\section{Related Work}

\head{Temporal Action Detection} or localization methods attempt to detect the start time and end time for action instances in untrimmed, long videos.  Standard benchmarks include THUMOS \cite{THUMOS14}, ActivityNet \cite{activitynet}, HACS \cite{zhao2019hacs}, etc.  All of them target a list of specified action classes and manually define the criteria for determining the start point and end point of each action, preventing annotations at scale.

Numerous methods have been developed for temporal action detection \cite{chao2018rethinking,dai2017temporal,gao2017cascaded,lin2018bsn,shou2017cdc,scnn_shou_wang_chang_cvpr16,alwassel2018action,lin2017single,zhao2017temporal,yuan2017temporal,long2019gaussian}. Notably, many of them contain a temporal proposal module which solves a binary classification problem analogous to foreground-background segmentation. ``Background" segments contain no pre-defined action classes.  However, many other generic events could appear in background segments, and segmenting generic events is the main focus in this paper. 

\head{Temporal Action Segmentation} \cite{ST-CNN,alayrac2017joint,kuehne2014language,farha2019ms,lei2018temporal,aakur2019perceptual,li2019weakly,richard2018action,huang2020improving} means labeling the action classes in every frames.  Some popular benchmarks are 50Salads \cite{stein2013combining}, GTEA \cite{lei2018temporal}, Breakfast \cite{kuehne2014language,Kuehne16end}, MERL Shopping \cite{singh2016multi}, etc.
Another task called \textbf{Temporal Action Parsing} was recently proposed in \cite{shao2020tapos}; parsing aims to detect the temporal boundaries for segmenting an action into sub-actions.  This is more closely related to our current work.  However, these annotations and methods are also developed for pre-defined action classes only, not generic boundaries.

\head{Shot Boundary Detection} is a classical task to detect shot transitions due to video editing such as scene cuts, fades/dissolves, and panning.
Some recent works are \cite{baraldi2015shot,gygli2018ridiculously,tang2018fast,shao2015shot,souvcek2019transnet}.  These shot boundaries are well-defined and an overcomplete set is easy to detect since the changes between shots are often significant.
In this paper, we also annotate and detect shot boundaries in our Kinetics-GEBD benchmark; however, the main novelty lies in event boundaries which are useful for breaking generic videos into semantically-coherent subparts.




\section{Definition of the GEBD Task}

\subsection{Task Definition}

GEBD localizes the moments where humans naturally perceive taxonomy-free event boundaries that break a longer event into shorter temporal segments.
To obtain ground truth annotations, we begin with the cognitive experiments' guideline \cite{tversky2013event} which achieved consistent boundaries marked by different annotators.
However, the cognitive experiments typically cover a limited number of scenarios in simple videos, e.g. a single actor, free of distractions from the event of interest.
We target diverse, natural human activity videos like Kinetics \cite{kay2017kinetics} which contain multiple actors, background distractions, different levels of detail in both space and time, etc.
Thus, there is more ambiguity about what are the event boundary positions.

\subsection{Principles for Designing Annotation Guideline}\label{principles}
To overcome these ambiguities in natural videos, we arrived at the following design principles throughout multiple iterations of improving annotation guidelines.

\head{(a) Detail in space: Focus on the dominant subject}. 
In order to avoid getting distracted by background events, annotators shall focus on the salient subject performing the event. The subject could be a person, a group, an object, or a collection of objects, depending on the video content.

\head{(b) Detail in time: Find event boundaries at ``1 level deeper'' granularity compared to the video-level event}. Given a video, it can be segmented at different temporal granularities. For example, the event boundaries of a long jump video could be 1) coarse: Long Jump starts / ends, or 2) intermediate: Long Jump is broken into running, jumping, and landing, or 3) fine: every foot step.  All variants are legitimate segmentations.  We embrace this ambiguity to a limited degree:  we instructed annotators to mark boundaries ``1 level deeper'' than the video-level event, and provided some examples but no precise definition of ``1 level''. Sometimes there is no one single video-level event; yet the merit of this principle is to ensure the segmented subparts are at the same level of granularity. This technique can be recursively applied to the segmented subparts when finer granularity is desired.  With this principle implemented, we find that humans can reliably agree on event boundaries without the need of a hand-crafted event boundary taxonomy.


\head{(c) Diversity of perception: Use multi-review}. 
Sometimes people have different interpretations of ``1 level deeper'' and go slightly deeper or coarser.
For example, in a video of two consecutive Long Jump instances, some might segment two instances of long jump, while others would segment the running and jumping units.
In practice, we consider both are correct and find that one video usually has at most 2-3 such possible variations due to the human perceiving differences rather than the ambiguity of task definition.
Thus, to capture such diversity, we assign 5 annotators for each video based the rule of thumb in user experience research.

\head{(d) Annotation format: Timestamps vs Time Ranges}.
The above principles clarify when to mark an event boundary.
The remaining question is marking where.
Following previous works, we can accommodate some ambiguity in ``where" during evaluation by varying an error tolerance threshold; more details can be found in Sec. \ref{sec:eval}.
We provide two options for marking an event boundary: 1) A single ``Timestamp'', typically used for instantaneous change (e.g. the moment when jumping begins in long jump).  2) A time ``Range'', typically used for short yet gradual change e.g. the interval between the end of landing and the start of standing up.  
More detailed can be found in Supp.

More details of our annotation guideline for Kinetics-GEBD (e.g. our own annotation interface, task rejection criteria, annotation format) can be found in Supp.

\subsection{Evaluation Protocol}\label{sec:eval}

As described in Sec. \ref{principles}, a boundary can be either a timestamp or a short range.
If it is a range, we represent it by its middle timestamp during evaluation.
Thus, our evaluation task is to measure the discrepancy between the detected timestamp and the ground truth timestamp, regardless of their types or semantic meanings.
To measure the discrepancy between timestamps, we follow previous works such as temporal parsing of an action instance \cite{shao2020tapos} and online detection of action start \cite{shou2018online} and use the Relative Distance (Rel.Dis.) measurement.
Inspired by the Intersection-over-Union measurement, Rel.Dis. is the error between the detected and ground truth timestamps, divided by the length of the corresponding whole action instance.
Given a fixed threshold for Rel.Dis., we can determine whether a detection is correct (i.e. $\leq$ threshold) or incorrect (i.e. $>$ threshold), and then compute precision, recall, F1 score for the whole dataset.
Note that duplicated detection for the same boundary is not allowed. Also, Each rater’s annotation is used separately. A detection result is compared against each rater’s annotation and the highest F1 score is treated as the final result.
We have also explored other metrics, e.g. the Global/Local Consistency
Error proposed in \cite{martin2004learning,martin2001database}, which are less appropriate here.
Detailed discussions can be found in Supp.



\section{Benchmark Creation: Kinetics-GEBD }

\subsection{Video Sources}

Our Kinetics-GEBD Train Set contains 20K videos randomly selected from Kinetics-400 Train Set \cite{kay2017kinetics}. Our Kinetics-GEBD Test Set contains another 20K videos randomly selected from Kinetics-400 Train Set.
Our Val Set contains all 20K videos in Kinetics-400 Val Set.

We rank all videos in Kinetics-400 Train set by video-level class. From this ordered list, we uniformly sample 20K videos as our Train Set and another 20K as our Test Set. Therefore, videos are chosen to have a similar distribution as Kinetics-400.

\subsection{Annotator Training}

To ramp up a new annotator, we provide a training curriculum consisting of a cascade of 5 training batches.
Each training batch contains 100 randomly sampled Kinetics videos with some reference annotations.
We make it clear to the annotator that different people may segment the same video in different ways, thus our provided annotations are only for reference.
Once a batch is done and before moving the annotator to the next batch, we will review its annotations for all 100 videos and provide specific feedback regarding errors made due to misunderstanding or misconduct of the guideline.
Overall, we do observe steady improvements over training batches for each new annotator.

\subsection{Quality Assurance}\label{sec:qa}

We present our detailed quality assurance mechanism in Supp.  Briefly, annotators were trained on 5 cascaded batches of 100 videos, with a QA mechanism before they worked on real jobs.  Typical issues early in training included  misunderstanding of the tool or guidelines, as well as annotating too much or too little detail.  Training videos were rated on a scale of 1 (good), 2 (minor errors like inaccurate timestamps), and 3 (bad, typically misunderstanding of guidelines). Raters progressed to real jobs when their average rating was deemed sufficient.
In practice, the performance of an annotator is satisfying and acceptable if its average rating is below 1.3.

\subsection{Common Characteristics of Boundary Causes}\label{sec:guideline}

\begin{figure}[h]
\begin{center}
\includegraphics[width=\linewidth]{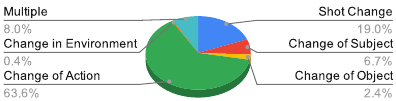}
\end{center}
\caption{Distribution of boundary causes on Kinetics-GEBD Val.}
\label{fig:dis_cause}
\end{figure}

Cognitive studies \cite{barker1955midwest} suggest that event boundaries can be characterized by several high-level causes.
Throughout our pilot annotation tasks for refining guideline, we confirmed such finding and arrived at the following high-level causes of event boundaries:
(1) \textbf{Change of Subject}: new subject appears or old subject disappears and such subject is dominant.
(2) \textbf{Change of Object of Interaction}: the subject starts to interact with a new object or finishes with an old object.
(3) \textbf{Change of Action}: an old action ends, or a new action starts.
Note that this characteristic includes when the subject changes physical direction (e.g. a runner suddenly changes direction) and when the same action is being performed multiple times (e.g. several consecutive push-up instances).
(4) \textbf{Change in Environment}: significant changes in color or brightness of the environment or the dominant subject (e.g. a light is turned on, illuminating a previously darker environment). 
Further, \textbf{Shot Change} boundaries are also common in Kinetics videos. Thus, we also annotate shot boundaries and the instructions can be found in Supp. In a video of multiple shots, the target granularity for event boundaries is 1 level deeper than the corresponding shot-level event.
Sometimes an event boundary might be due to \textbf{Multiple} coupled causes or \textbf{Others}.
As the distribution shown in Fig.~\ref{fig:dis_cause}: \textbf{Others} is negligible; \textbf{Change of Action} is the most common cause. Note that the actions leading to boundaries in our dataset are \textit{much more generic and diverse} than the pre-defined taxonomies in the current CV action datasets.


\subsection{Annotation Results Summary and Analysis}

\head{Annotation capacity}. 
In total, around 40 qualified annotators were trained to annotate our Kinetics-GEBD.
The average speed is around 5mins per video per annotator.

\head{Statistics of \#annotations received}.
Recall that each video is annotated by 5 annotators.
Annotators can reject a video due to the reasons stated in Supp.
Table \ref{table:num_videos_per_mr} shows that most videos receive all 5 annotations without rejection.

\begin{table}[h]\footnotesize
\begin{center}
\begin{tabular}{ccccccc}
\#Annotations & 0 & 1 & 2 & 3 & 4 & 5  \\ \hline 
\#videos & 101 & 141 & 203 & 342 & 805 & 18166 \\
Per. (\%) & 0.51 & 0.71 & 1.03 & 1.73 & 4.07 & 91.94
\end{tabular}
\end{center}
\caption{For our Kinetics-GEBD Val set, \#annotations received per video vs. \#videos and its percentage .}
\label{table:num_videos_per_mr}
\end{table}

\head{The extent of consensus for GEBD annotation}.
Given the construction of the dataset, a natural question is ``how consistent are the annotations?''.  Adopting the protocol in Sec.~\ref{sec:eval}, for the same video, we treat one annotation as ground truth and another annotation as detection result.  Since we expect consistent annotations to have very close boundaries in time, we do not use relative distance; instead, we evaluate F1 score based on the absolute distance between two boundaries, varying the threshold from 0.2s to 1s with a step of 0.2s, and calculate the average F1 score.
By averaging the F1 score over all pairs of annotations for the same video, we can obtain its consistency score.
If all raters make very similar annotations, the consistency score will be high i.e. towards 1; otherwise low i.e. towards 0.

\begin{figure}[b]
\begin{center}
\includegraphics[width=\linewidth]{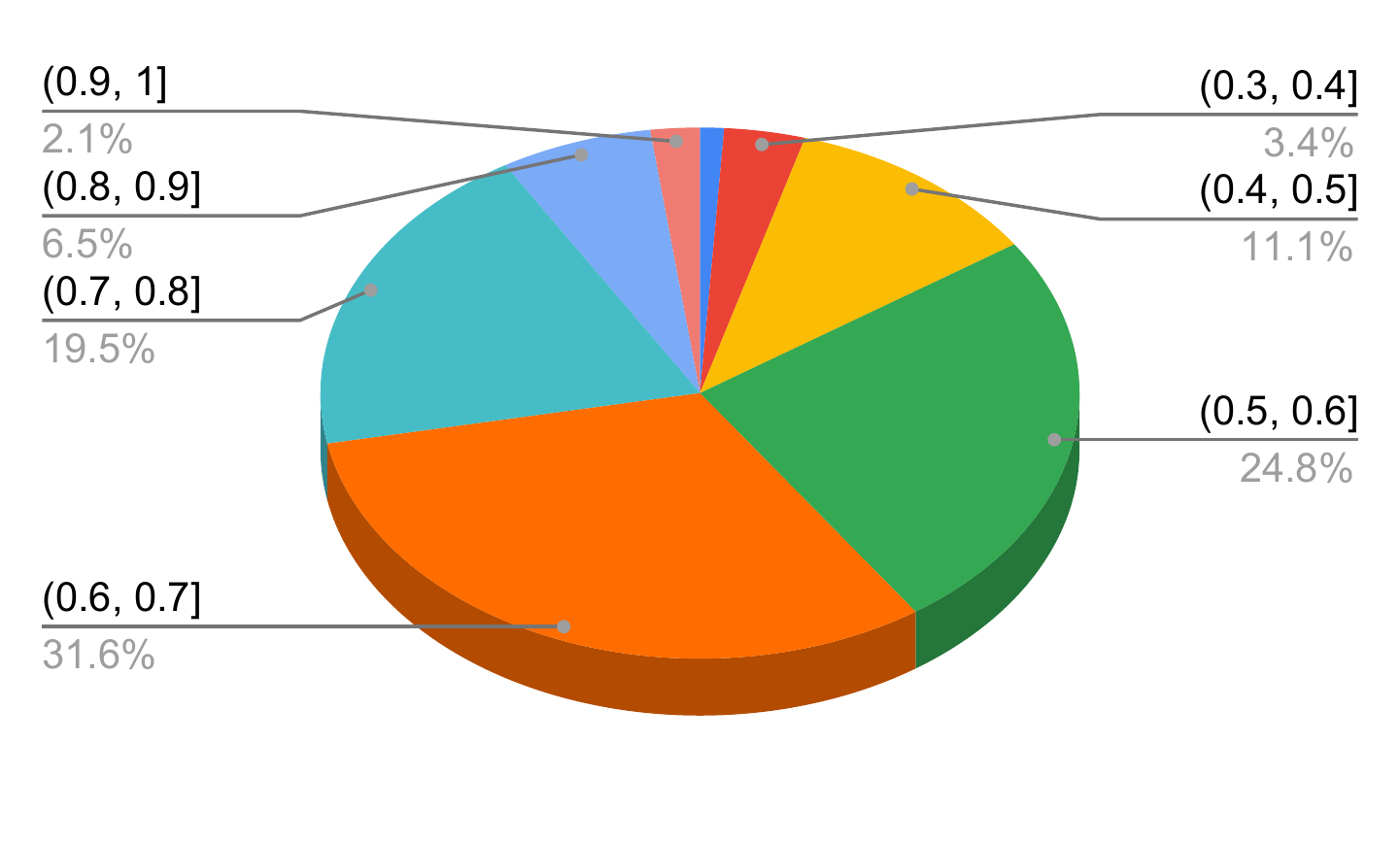}
\end{center}
\vspace{-3em}
\caption{The number of videos percentage (below line) for each range of the consistency score (above line) on our Kinetics-GEBD Val set when the video is not rejected by any annotators.}
\label{fig:num_videos_per_f1}
\end{figure}

Fig.~\ref{fig:num_videos_per_f1} shows that the majority of videos have consistency scores higher than 0.5.
This indicates that given our designed task definition and annotation guideline, humans are able to reach decent degree of consensus, taking into account the factors that (1) often due to different human perception manners, a video can have multiple correct segmentations, and (2) sometimes annotators make mistakes.

To understand how the frequency of annotation mistakes (i.e. annotation quality) correlates with the consistency score, in Table~\ref{table:manual_vs_f1}, we randomly sample 5 non-rejection videos for each consistency score range and conduct manual auditing according to the protocol in Sec.~\ref{sec:qa} to get the average rating for each range.
As the consistency score becomes low, the rating gets worse.  Recall that the cutoff for the rating to determine qualified annotators is 1.3, which corresponds to 0.5 consistency score here.

\begin{table}[h]\footnotesize
\begin{center}
\begin{tabular}{cccccc}
Consistency & (0.4,0.5] & (0.5,0.6] & (0.6,0.7] & (0.7,0.8] & (0.8,1]  \\ \hline
Rating & 1.4 & 1.24 & 1.20  & 1.16 & 1.04
\end{tabular}
\end{center}
\caption{Average audit rating vs. average F1 consistency score on our Kinetics-GEBD Val set.}
\label{table:manual_vs_f1}
\end{table}


\subsection{Post-processing for the Annotations}
\label{sec:post-processing}
Given the raw annotations, we conduct the following steps to construct our Kinetics-GEBD benchmark.
(1) To ensure annotation quality and remove very ambiguous videos, we exclude videos that have lower than 0.3 consistency score.
(2) To capture the diversity of human perception, we only keep videos that receive at least 3 annotations.
During evaluation, the detection is compared against each ground truth annotation and the highest F1 score is treated as the final result.
(3) For each annotation, if two boundaries are very close (i.e. less than 0.1s), we merge them into one. Note that this includes the case that one Timestamp boundary falls into a Range, or one Range boundary overlaps with another Range boundary.  We remove any boundaries from the initial and final 0.3s of each video. More details of our motivation for annotation post-processing could be found in Supp.

\subsection{Statistics}
\begin{figure}[h]
    \subfloat[]{\includegraphics[width=\linewidth]{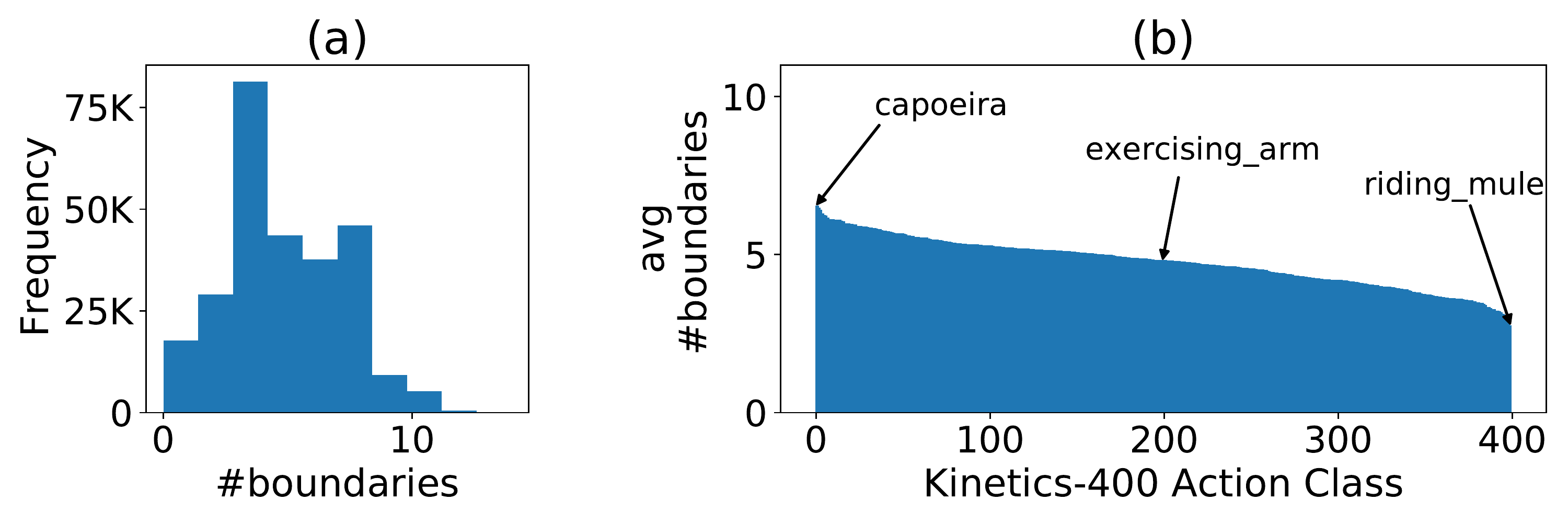}} \vspace{-0.5cm}\\ 
    \subfloat[]{\includegraphics[width=\linewidth]{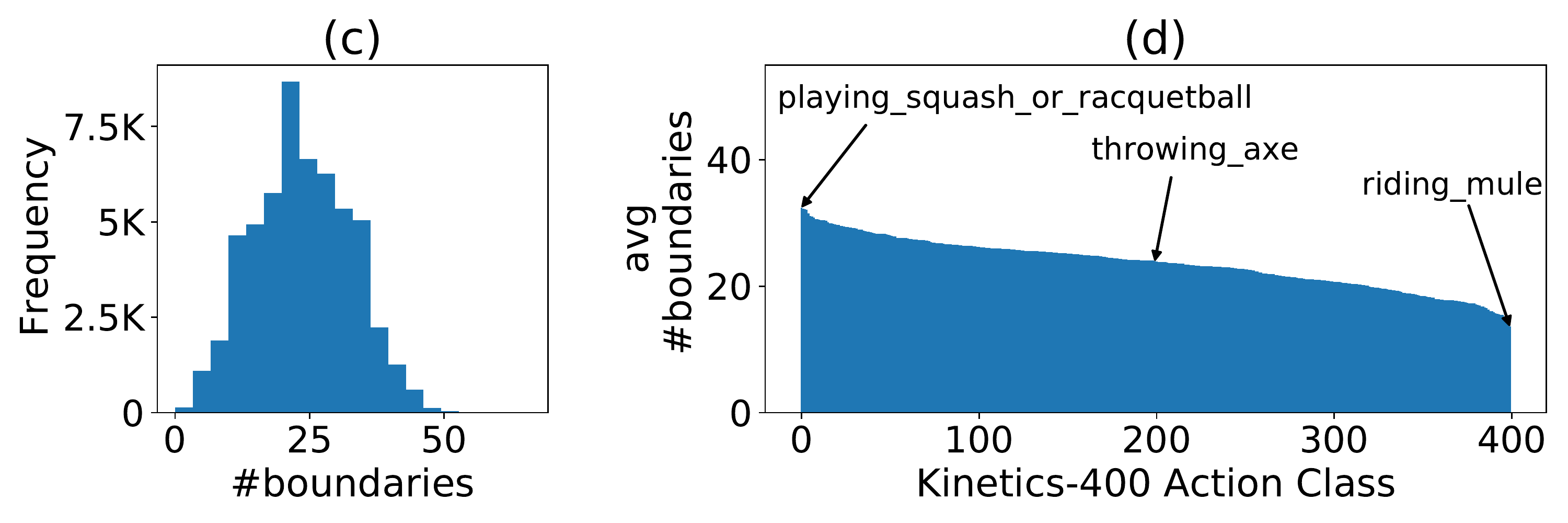}} \vspace{-0.5cm}\\ 
    \subfloat[]{\includegraphics[width=\linewidth]{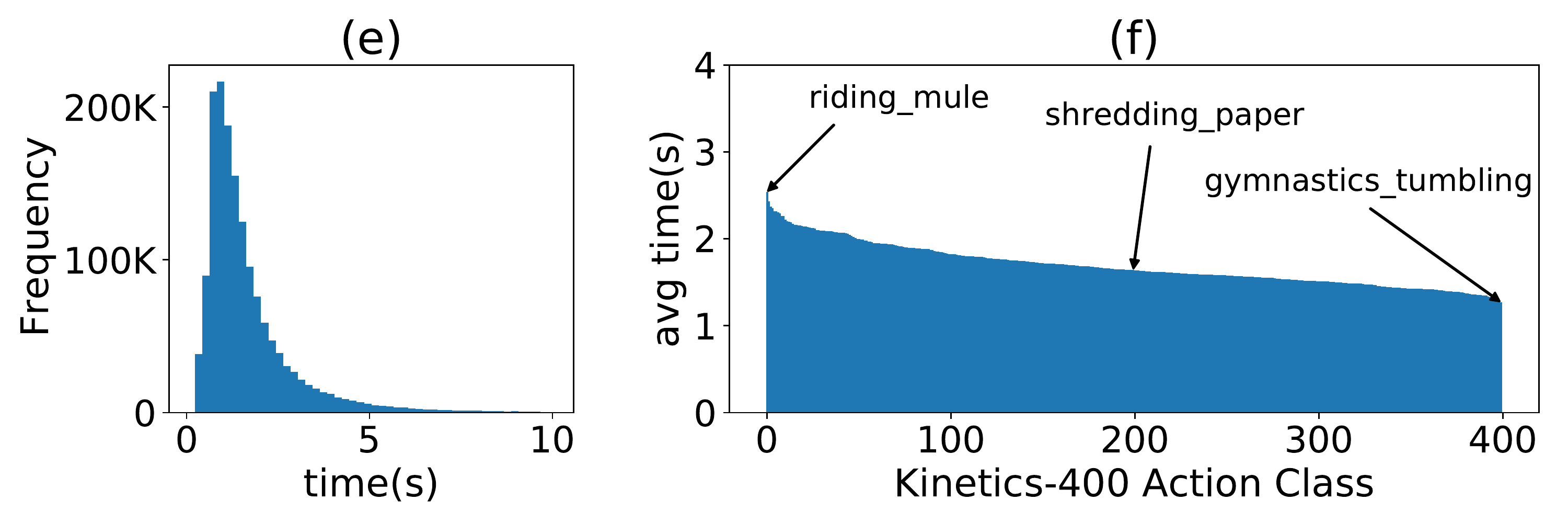}} \vspace{-0.5cm}
    \caption{Statistics on Kinetics-GEBD. \#boundaries per video per annotation: (a) distribution (b) average over each Kinetics class and then sorted by class; \#boundaries per video: (c) distribution (d) average over each Kinetics class and then sorted by class; duration per segment: (e) distribution (f) average over each Kinetics class and then sorted by class.}
    \label{fig:GEBD-stat}
\end{figure}
For the raw Kinetics-GEBD annotation, the average number of boundaries per video per annotation is 5.48 (std dev 2.76, range [1, 33]).
The average time between boundaries is 1.47s (std dev 1.24, range [0,10.01]).
The number and average length of the time-range boundaries is 265K and 0.71s.
The number of timestamp-only boundaries is 1232K. 

For the Kinetics-GEBD benchmark (after post-processing raw annotations), the average number of boundaries per video per annotation is 4.77 (std dev 2.24, range [0,14], distribution plot as Fig.~\ref{fig:GEBD-stat}(a)).
The average time between boundaries is 1.65s (std dev 1.25, range [0.023, 10.08], distribution plot as Fig.~\ref{fig:GEBD-stat}(e)). 

Furthermore, the left column of Fig.~\ref{fig:GEBD-stat} shows the distribution of \#boundaries per video per annotation, \#boundaries per video and duration per segment, respectively. 
To show how these compared on the base Kinetics-400
classes, we rank all Kinetics classes from high to low and highlight 3 classes, as shown in the right column of Fig.~\ref{fig:GEBD-stat}.

\section{Experimental Results of GEBD Methods}\label{sec:exp}

\subsection{Dataset}

In addition to our Kinetics-GEBD, we also experiment on the recent TAPOS dataset \cite{shao2020tapos} containing Olympics sport videos with 21 actions.
The training set contains 13,094 action instances and the validation set contains 1,790 action instances. The authors manually defined how to break each action into sub-actions during annotation.
While not taxonomy-free, the TAPOS boundaries between sub-actions are analogous to GEBD action boundaries.  Thus, we can re-purpose TAPOS for our GEBD task by trimming each action instance with its action label hidden  (can be as long as 5mins) and conducting GEBD on each action instance.
Note that in TAPOS only 1 rater's annotation has been released and thereby used as ground truth.

\begin{table*}[t]
\begin{tabular}{ccccccccccccc}                                                                                            &                \\ \hline
                                               \multicolumn{2}{c|}{Rel.Dis. threshold} & 0.05           & 0.1            & 0.15           & 0.2            & 0.25           & 0.3            & 0.35           & 0.4            & 0.45           & \multicolumn{1}{c|}{0.5}            & avg            \\ \hline
\multicolumn{1}{c|}{\multirow{3}{*}{Unsuper.}} & \multicolumn{1}{c|}{SceneDetect}    & 0.035          & 0.045          & 0.047          & 0.051          & 0.053          & 0.054          & 0.055          & 0.056          & 0.057          & \multicolumn{1}{c|}{0.058}          & 0.051          \\
\multicolumn{1}{c|}{}                          & \multicolumn{1}{c|}{PA - Random}   & 0.158          & 0.233          & 0.273          & 0.310          & 0.331          & 0.347          & 0.357          & 0.369          & 0.376          & \multicolumn{1}{c|}{0.384}          &    0.314        \\

\multicolumn{1}{c|}{}                          &
\multicolumn{1}{c|}{PA}      & \textbf{0.360} & \textbf{0.459} & \textbf{0.507} & \textbf{0.543} & \textbf{0.567} & \textbf{0.579} & \textbf{0.592} & \textbf{0.601} & \textbf{0.609} & \multicolumn{1}{c|}{\textbf{0.615}} & \textbf{0.543} \\
\hline

\multicolumn{1}{c|}{\multirow{5}{*}{Super.}}   & \multicolumn{1}{c|}{ISBA}           & 0.106          & 0.170          & 0.227          & 0.265          & 0.298          & 0.326          & 0.348          & 0.369          & 0.382          & \multicolumn{1}{c|}{0.396}          & 0.302          \\
\multicolumn{1}{c|}{}                  & \multicolumn{1}{c|}{TCN}            & 0.237          & 0.312          & 0.331          & 0.339          & 0.342          & 0.344          & 0.347          & 0.348          & 0.348          & \multicolumn{1}{c|}{0.348}          & 0.330                  \\
\multicolumn{1}{c|}{}                          & \multicolumn{1}{c|}{CTM}            & 0.244          & 0.312          & 0.336          & 0.351          & 0.361          & 0.369          & 0.374          & 0.381          & 0.383          & \multicolumn{1}{c|}{0.385}          & 0.350          \\
\multicolumn{1}{c|}{}                          & \multicolumn{1}{c|}{TransParser}    & 0.289          & 0.381          & 0.435          & 0.475          & 0.500       & 0.514          & 0.527          & 0.534          & 0.540          & \multicolumn{1}{c|}{0.545}          & 0.474          \\
\multicolumn{1}{c|}{}                          & \multicolumn{1}{c|}{PC}    & \textbf{0.522}          & \textbf{0.595}          & \textbf{0.628}          & \textbf{0.646}          & \textbf{0.659}       & \textbf{0.665}          & \textbf{0.671}          & \textbf{0.676}          & \textbf{0.679}          & \multicolumn{1}{c|}{\textbf{0.683}}          & \textbf{0.642}          \\ \hline
\end{tabular}
\vspace{1.5em}
\caption{F1 results on TAPOS for various supervised and unsuperivsed GEBD methods.}
\label{table:Kin_TAPOS}
\end{table*}

\begin{table*}[t]
\begin{center}
\begin{tabular}{ccccccccccccc}                                                                                            &                \\ \hline
                                               \multicolumn{2}{c|}{Rel.Dis. threshold} & 0.05           & 0.1            & 0.15           & 0.2            & 0.25           & 0.3            & 0.35           & 0.4            & 0.45           & \multicolumn{1}{c|}{0.5}            & avg            \\ \hline
\multicolumn{1}{c|}{\multirow{3}{*}{Unsuper.}} & \multicolumn{1}{c|}{SceneDetect}    & 0.275 & 0.300          & 0.312          & 0.319          & 0.324          & 0.327          & 0.330          & 0.332          & 0.334          & \multicolumn{1}{c|}{0.335}          & 0.318          \\
\multicolumn{1}{c|}{}                          & \multicolumn{1}{c|}{PA - Random}    & 0.336          & 0.435          & 0.484         & 0.512          & 0.529          & 0.541          & 0.548          & 0.554          & 0.558          & \multicolumn{1}{c|}{0.561}          & 0.506          \\

\multicolumn{1}{c|}{}                          & \multicolumn{1}{c|}{PA}      & \textbf{0.396}          & \textbf{0.488} & \textbf{0.520} & \textbf{0.534} & \textbf{0.544} & \textbf{0.550} & \textbf{0.555} & \textbf{0.558} & \textbf{0.561} & \multicolumn{1}{c|}{\textbf{0.564}} & \textbf{0.527}
\\ \hline

\multicolumn{1}{c|}{\multirow{5}{*}{Super.}}   & \multicolumn{1}{c|}{BMN}            & 0.186        & 0.204 & 0.213 & 0.220 & 0.226 & 0.230 & 0.233 & 0.237 & 0.239 & \multicolumn{1}{c|}{0.241}          &  0.223   \\
\multicolumn{1}{c|}{}   & \multicolumn{1}{c|}{BMN-StartEnd}            & 0.491 & 0.589 & 0.627 & 0.648 & 0.660 & 0.668 & 0.674 & 0.678 & 0.681 & \multicolumn{1}{c|}{0.683}          &  0.640  \\
\multicolumn{1}{c|}{}                          & \multicolumn{1}{c|}{TCN-TAPOS}            & 0.464  & 0.560  & 0.602  & 0.628  & 0.645  & 0.659  & 0.669  & 0.676  & 0.682  & \multicolumn{1}{c|}{0.687}          &   0.627
\\ 
\multicolumn{1}{c|}{}                          & \multicolumn{1}{c|}{TCN}            & 0.588 & 0.657 & 0.679 & 0.691 & 0.698 & 0.703 & 0.706 & 0.708 & 0.710 & \multicolumn{1}{c|}{0.712}          &   0.685
\\
\multicolumn{1}{c|}{}                          & \multicolumn{1}{c|}{PC}            & \textbf{0.625}  & \textbf{0.758}  & \textbf{0.804}  & \textbf{0.829}  & \textbf{0.844}  & \textbf{0.853}  & \textbf{0.859}  & \textbf{0.864}  & \textbf{0.867}  & \multicolumn{1}{c|}{\textbf{0.870}}          &   \textbf{0.817}
\\ \hline
\end{tabular}
\end{center}
\caption{F1 results on Kinetics-GEBD for various supervised and unsuperivsed GEBD methods.}
\label{table:Kin_GEBD}
\end{table*}

\subsection{Supervised Methods for GEBD}

We directly quote the results of supervised methods from \cite{shao2020tapos} on TAPOS (i.e. the below \textbf{\#1-3}). Since \cite{shao2020tapos} has not published codes, we implement the below \textbf{\#4-6} methods by ourselves on our Kinetics-GEBD:

\head{\#1.} Temporal parsing model: \textbf{TransParser} \cite{shao2020tapos} proposes a pattern miner trained with a local loss based on the sub-action boundary supervision and a global loss trained with the action instance label supervision.

\head{\#2.} Temporal action segmentation models: 
Connectionist Temporal Modeling (\textbf{CTM}) \cite{huang2016connectionist} and
Iterative Soft Boundary Assignment (\textbf{ISBA}) \cite{Ding2018WeaklySupervisedAS} are supervised by the order of occurrence of a set of pre-defined sub-actions.

\head{\#3.} Action boundary detection model: Temporal Convolution Network (\textbf{TCN}) \cite{ST-CNN,lin2018bsn} trains a binary classifier to distinguish the frames around boundaries against other frames.

\head{\#4.} \textbf{Pairwise boundary Classifier (PC)}: 
At each candidate boundary position time $t$, we use the same backbone network to extract a feature pair: the average feature of frames before and the average feature of frames after $t$.
We conduct global pooling over space for each feature and then we concatenate these two paired features together as the input to a linear binary classifier, which is trained to predict the probability of time $t$ is a boundary.
\textbf{PC} is trained end-to-end to fine-tune the backbone network pre-trained on ImageNet; training with the backbone fixed does not converge.
We watershed the probability sequence to obtain internals above 0.5.
Each internal's center is treated as an event boundary.

\head{\#5.} Temporal action proposal model: to understand how well an class-agnostic action boundary proposal model can detect generic event boundaries, we train a BMN model \cite{lin2019bmn} on THUMOS'14 \cite{THUMOS14} and test it on Kinetics-GEBD to generate action proposals.
We denote \textbf{BMN} as treating both the start and end of each action proposal as event boundary.
Alternatively, since one intermediate step in BMN is to evaluate two probability scores of respectively being action start and end, we watershed each probability sequence to obtain internals above 0.5 and treat the center of each internal as an event boundary. We take the union of all these centers and denote this method as \textbf{BMN-StartEnd}.

\head{\#6.} Cross-dataset GEBD method \textbf{TCN-TAPOS}: to confirm the need of Kinetics-GEBD which is more challenging than TAPOS, we conduct testing on Kinetics-GEBD using the TCN model trained on TAPOS.

\subsection{Unsupervised Methods for GEBD}

This direction is intriguing because it can potentially handle any kind of events, without the need to annotate a large amount of event boundary labels.

\head{\#1.} \textbf{SceneDetect}\footnote{https://github.com/Breakthrough/PySceneDetect}: an online popular library for detecting classical shot changes.

\head{\#2.} \textbf{PA - Random}: we randomly swap the detection results of the below \textbf{PA} method among all videos. The position of each boundary is mapped to the new video with its relative position in the original video unchanged. 

\head{\#3.} \textbf{PredictAbility (PA)}:
Event Segmentation Theory indicates that the moment people perceive event boundary is where future activity is least predictable \cite{kurby2008segmentation,reynolds2007computational,zacks2007event}. 
This motivates us to develop a \textbf{PA}-based method which first 1) computationally assesses the predictability score over time and then 2) locates the event boundaries by detecting the local minima of the predictability sequence.
 

1) \textit{Predictability Assessment}\label{sec:pa}:
To quantify the predictability at time $t$, we compute the average feature of frames preceding and the average feature of frames succeeding $t$.
Then, we compute their squared L2 norm feature distance to obtain the inverse predictability $\phi \left ( t \right )$; lower distance implies greater predictability.

2) \textit{Boundaries from Predictability}\label{sec:bdydet}:
Given $\phi \left ( t \right )$, a natural method is to propose temporal boundaries at the local maxima of $\phi$. This is similar to the classical blob detection problem, and thus we apply the classical Laplacian of Gaussian (LoG) filter \cite{lindeberg1998feature} to our 1D temporal problem.
We apply the 1D LoG filter to compute $L(t) = \mathrm{LoG}(\phi \left ( t \right ))$, and compute its derivative $L' \left ( t \right )$.  We detect temporal boundaries at the negative-to-positive zero-crossings of $L'$, which correspond to local maxima of $\phi$.


\subsection{Implementation Details}\label{sec:details}

The following settings are used for all experiments conducted by ourselves unless explicitly specified otherwise:
2 GP100 NVIDIA cards are used.
For each video, we sample 1 frame for every 3 frames.
The inputs are RGB images resized to 224x224.
To make fair comparisons, all models implemented by ourselves, i.e. \textbf{PC}, \textbf{TCN}, \textbf{TCN-TAPOS}, \textbf{PA}, \textbf{BMN}, \textbf{BMN-StartEnd}, build on ResNet-50 \cite{He_2016_CVPR} backbone.
\textbf{PC} is trained end-to-end while others simply use the off-the-shelf ImageNet pretrained feature.
Our \textbf{PC}, \textbf{TCN}, \textbf{TCN-TAPOS} and \textbf{PA} all use 5 frames before and 5 frames after a candidate boundary as the model input.
For \textbf{PA}, we tune the sigma in the LoG filter on the Train set and set it to 15.
During evaluation, we follow TAPOS \cite{shao2020tapos} to vary the Relative Distance (Rel.Dis.) threshold indicated in Sec. \ref{sec:eval} from 5\% to 50\% with a step of 5\%.

\subsection{Results Comparisons}\label{sec:exp-tapos}

\head{\underline{TAPOS}} val set F1 results are shown in Table~\ref{table:Kin_TAPOS}. 
Detailed results of precision and recall are in Supp.
The predictability-based \textbf{PA} method is clearly much better than the random guess.
It is quite encouraging to see that our unsupervised method \textbf{PA} even outperforms all previous supervised methods i.e. ISBA, TCN, CTM, TransParser.
\textbf{SceneDetect} achieves high precision while quite low recall because it only fires at the very salient boundaries.

\head{\underline{Kinetics-GEBD}} val set F1 results are shown in Table~\ref{table:Kin_GEBD}.
Detailed results of precision and recall are in Supp.
Among unsupervised methods, \textbf{PA} is clearly better than shot change detection method \textbf{SceneDetect} and the random guess \textbf{PA - Random}. 
Comparing \textbf{PA} with the supervised method \textbf{TCN} which also uses the same fixed backbone feature, the gap is not large, indicating that un-supervised or semi-supervised GEBD methods are worthwhile researching in the future.
\textbf{PC} clearly outperforms others, indicating that the event boundaries cannot be comprehensively represented by off-the-shelf feature while can be better learned by the backbone.
For the class-agnostic action proposal methods, directly detecting action proposals (i.e. \textbf{BMN}) is not a good GEBD approach but accessing the probability of being boundary (i.e. \textbf{BMN-StartEnd}) is effective.
\textbf{BMN-StartEnd} is still worse than \textbf{PC} due to only detecting action change boundaries while ignoring other generic event boundaries like subject change.
For the similar reason, on Kinetics-GEBD, a GEBD model trained on TAPOS (i.e. \textbf{TCN-TAPOS}) underperforms the same model directly trained on Kinetics-GEBD (i.e. \textbf{TCN}).
These again confirm the challenging nature of generic event boundaries and the need of our new benchmark Kinetics-GEBD.

\section{Applications of Video Event Boundaries}

\begin{figure}[h]
\begin{center}
\includegraphics[width=0.9\linewidth]{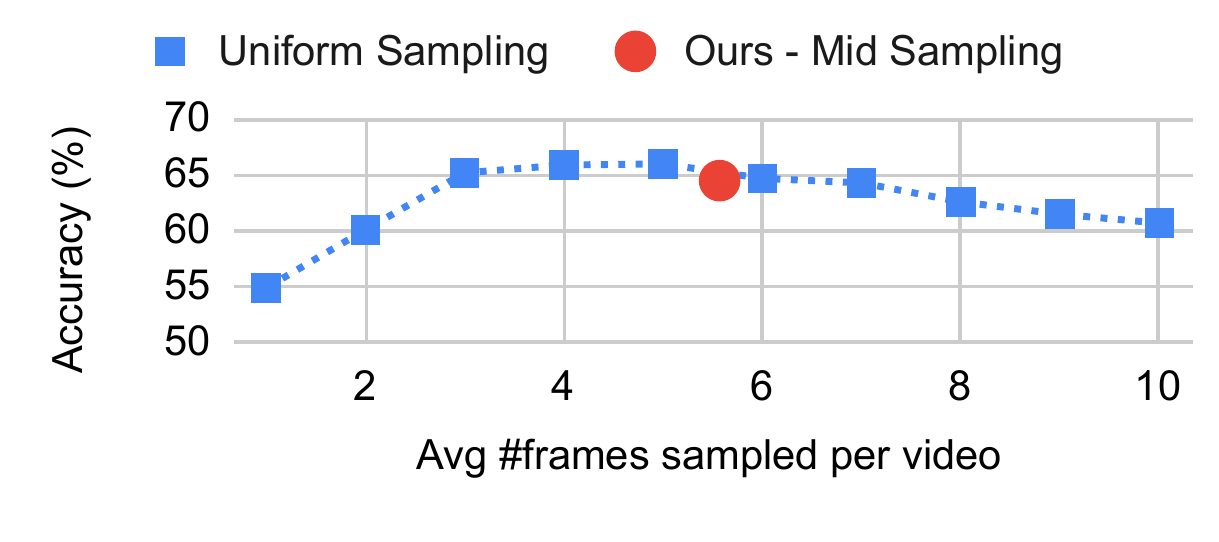}
\end{center}
\vspace{-1.5em}
\caption{To classify a video, it is difficult to tell what is the optimal number of frames for uniform sampling. Our event boundaries provide cue about how many frames shall be sampled.}
\label{fig:Kin_Cls}
\end{figure}

\subsection{Video-level classification}

We test the classification accuracy on videos that receive at least 3 annotations in Kinetics-GEBD Val set.
We use the public implementation\footnote{https://github.com/mit-han-lab/temporal-shift-module} of the TSN \cite{TSN} model which uniformly samples $K$ frames, applies ResNet-50 backbone on each frame, and finally average the predictions to get the video-level prediction.
Fig.~\ref{fig:Kin_Cls} shows that the video-level classification accuracy for uniform sampling (blue curve) increases and then decreases as $K$ varies from 1 to 10.
Thus, give a video, how can we determine $K$?

Despite GEBD is not designed to select discriminative frames, \textit{our boundaries provide cue about how to set $K$ in uniform sampling in order to achieve high classification accuracy}
Based on our annotated boundaries, we can break the video into segments and each segment might only need one frame to be sampled.
To validate this hypothesis, we sample the middle frame of each segment. Fig.~\ref{fig:Kin_Cls} shows that this (the red dot) uses in average 5.5 frames per video and achieves accuracy close to the best achieved by uniform sampling.  
This is useful in practice when the video content is diverse and thus we do not know what is the best $K$.

In addition, we find that sampling the middle frame (64.4\% accuracy) outperforms sampling boundary frames (63.0\% accuracy).
This implies that GEBD helps identify less-discriminative frames (\textit{the frames at boundaries are less discriminative}), and is consistent with the cognitive finding that boundaries have less predictive power.

\subsection{Video summarization}

Our temporal boundaries provide a natural way to select keyframes for video summarization.
We conduct the following two user study tasks to compare \textbf{Ours} (sample the middle frame of subparts) and \textbf{Uniform} (uniformly sample the same number of frames as Ours).
In Task 1, we randomly sample videos from Kinetics-GEBD Val.
In Task 2, we select the videos that the frame distance between \textbf{Ours} and \textbf{Uniform} are the largest.
Each task involves around 200-250 videos.

For each video in both task, 20 users are asked ``which set of keyframes better summarize the video comprehensively?'' and shall vote one out of three options: (1) Set 1 is better; (2) Set 2 is better; (3) Tie (both good/bad summarization).
Table~\ref{table:sum} shows the percentage of different options winning at vote-level and at the video-level (e.g. out of 20 votes for the same video, if \#votes for (1) is the highest, Set 1 wins).
We can see that for random samples \textbf{Ours} is clearly better than \textbf{Uniform} and for samples of large disparity \textbf{Ours} significantly outperforms \textbf{Uniform}.

\begin{table}[h]\footnotesize
\begin{center}
\begin{tabular}{cc|ccc}
\multicolumn{2}{c|}{Percentage (\%)}                                & Uniform & Ours & Tie  \\ \hline \hline

\multicolumn{1}{c|}{\multirow{2}{*}{Task 1: random samples}}  & Vote-level  & 33.9    & \textbf{40.9} & 25.1 \\ \cline{2-5} 
\multicolumn{1}{c|}{}                                & Video-level & 38.3    & \textbf{43.7} & 17.8 \\ \hline

\multicolumn{1}{c|}{\multirow{2}{*}{Task 2: large disparity}} & Vote-level  & 12.6    & \textbf{73.0} & 14.3 \\ \cline{2-5} 
\multicolumn{1}{c|}{}                                & Video-level & 6.0     & \textbf{90.0} & 4.0 \\ \hline

\end{tabular}
\end{center}
\vspace{-1em}
\caption{User study results for video summarization.}
\label{table:sum}
\end{table}

\section{Conclusion and Future Work}

In this paper, we have introduced the new task of GEBD and resolved ambiguities in the annotation process.
A new benchmark, Kinetics-GEBD, has been created along with novel designs for annotation guidelines and quality assurance.
we benchmark supervised and un-supervised GEBD approaches on the TAPOS dataset and our Kinetics-GEBD, together with method design explorations that suggest future directions.
We also showed value for GEBD in downstream applications.

We believe our work is an important stepping stone towards long-form video understanding and hope it will enable future work in learning based on temporal event structure.
In the future, we plan to address scene changes which usually happen in much longer videos (e.g. move from kitchen to bathroom in 30mins long ADL \cite{pirsiavash2012detecting} videos, move from street to restaurant in hours long UT-Ego \cite{lee2012discovering} videos).


\section{Acknowledgement}

We thank Jitendra Malik for the insightful guidance.
We appreciate the great support in annotation from our Product Data Operations team and Annotation Tooling team.
Mike Shou and Stan Lei are supported by the National Research Foundation Singapore under its NRFF award NRF-NRFF13-2021-0008.

\section{Supp: More Details of the GEBD Task}

\subsection{Other Candidate Evaluation Metrics}

We made attempts to explore adapting the Global/Local Consistency Error metric, which was proposed in \cite{martin2004learning,martin2001database} to evaluate boundary segmentation of 2D image, from 2D spatial to 1D temporal for our task. 
This metric is designed for the scenario that we do not impose penalty if the detection and ground truth are of different granularity i.e. one is a fine-grained, detailed segmentation of another one.
During our experiments, we found the variance of this metric between different detection results, regardless of their qualities, is often quite small.
This is because some incorrect boundaries made by one detection can be mistakenly considered as a correct, finer segmentation of a ground truth segmentation.
Thus, during annotators training we cannot use this metric to quickly identify bad annotations to provide feedback to annotators, and during benchmarking we are not able to always effectively distinguish the bad detection against good detection.
Consequently, we do not end up with using this metric.
In aligning with this, our task is also designed to target specific temporal granularity i.e. 1 level deeper in semantics compared to the video-level event.

\subsection{Discussion of Much Longer Videos and Scene Change Boundaries: Future Work}

In this paper, since the video source for our benchmark is Kinetics \cite{kay2017kinetics}, our target boundaries are event boundaries.
This is because each video corresponds to one single, dominant event at the whole video level.
Its videos usually are of 10s long, which is longer than the effective input clip duration of the mainstream video models i.e. 1-2s and thus still indeed presents the challenge of whole, long-form video modeling.
For such 10s Kinetics videos, two types of boundaries are prevalent for our GEBD task: 1) shot change boundary due to editing, and 2) event boundary which breaks an event into temporal units.

In addition to Kinetics, initially we have also considered other data sources e.g. longer videos such as 30mins long videos in ADL \cite{pirsiavash2012detecting} and hours long videos in UT-Ego \cite{lee2012discovering}.
We conducted preliminary studies and got two observations:
(1) Compared to Kinetics-GEBD, these long videos contain too many event boundaries to be annotated completely.
(2) Recall the principle stated in the main paper that targeting event boundaries at 1 level deeper in compared to the whole video-level event.
When adapting this principle to these very long videos, the temporal boundaries often happen at the moments of scene/situational changes (e.g. the moment moving from kitchen to bathroom in ADL \cite{pirsiavash2012detecting}, the moment changing from street to restaurant in UT-Ego \cite{lee2012discovering}).
Thus, such videos are more suitable for studying scene/situational boundary detection.
In practice, if need to detect event boundaries on such very long videos, the models developed based on our Kinetics-GEBD can also be applied, and even be applied recursively to deal with event boundaries at different temporal granularity.

In summary, we believe both scenarios are important. In this paper, we focus on the scenario that the video is at around 10s timescale and the whole video presents one event. The event can be segmented into temporal units based on shot boundaries and event boundaries which are the main focus in this paper and benchmark.
In the future, it would be interesting to specifically annotate and build benchmark for the scenario that involves much longer videos and focuses on scene change boundaries, and we will not need to annotate event boundaries again.

\section{Supp: More Details of the Kinetics-GEBD Annotation}

\subsection{Guideline: A Mixture of Descriptions and Visual Examples}

A very useful practice we learned to eliminate ambiguity is that when designing annotation guideline, use both visual examples and textual descriptions. Examples are straightforward to illustrate the requirements but they are too specific and cannot cover every cases. Descriptions can cover the generic requirements but often are too abstract to be understood clearly. 

\subsection{The annotation format - Timestamps vs Time Ranges for Kinetics-GEBD}

The term ``Relatively'' has been used multiple times in the cognitive survey \cite{tversky2013event} and turns out to be a good practice.
For examples, Kinetics \cite{Kinetics} videos are usually 10s long and each video corresponds to one event. Hence, we ask the annotators to be very cautious whenever marking a Range if its possible duration is less than 0.3s. Note that on Kinetics, usually a boundary's duration will not exceed 2s otherwise it is likely a meaningful temporal unit rather than just a transition between units.
Therefore, we request the annotators to be very cautious whenever they want to mark a range longer than 2s.

\subsection{Rejection Scenarios during Annotation}

We provide the option for annotators to reject a video when 1) the video does not have significant change over time and thus does not contain temporal boundary; 2) the video contains violating content e.g. nudity, graphic violence; 3) the video has been edited to play at extreme speed; 4) the video cannot be understood e.g. too blurry all the time; 5) the video contains too many boundaries to be annotated accurately i.e. more than 20 boundaries for a 10s long video in Kinetics-GEBD.

\subsection{Boundary Causes: Shot Change}

Shot change boundaries correspond to the changes caused by video editing or rapid camera behaviors. (1) Some shot change boundaries are always sudden and shall be annotated using one single \textbf{Timestamp: Cut, Change from/to slow motion, Change from/to fast motion}. (2) Some shot change boundaries are always gradual and shall be annotated using a short \textbf{Range: Change due to Panning, Change due to Zooming, Change due to Fade/Dissolve/Gradual}. 

Note that for Panning and Zooming, the goal is not to simply mark whenever the camera pans or zooms; these are just characteristics of some common shot change boundaries; the marked boundary has to be at the moment of a change that connects two temporal units and the succeeding unit brings in new information, and meanwhile such change is also a panning effect.
For example, during the running period of long jump in Fig.~\ref{fig:teaser} in the main paper, after the scene cut (i.e. the period corresponding to the second frame), despite the camera has been panning, while the dominant person is in the center all the time and there is no action change or other new information. Therefore, there is no temporal event boundary shall be marked during the period corresponding to the second frame.

\subsection{Event Class vs. High-Level Cause}
In contrast to high-level causes, we consider generic event classes are low-level and capture specific details, which is similar to classes in conventional action datasets.
Importantly, such a handful of high-level causes 1) are not exclusive - the dropdown does have the \textbf{Others (N/A) option} for a boundary cannot be characterized by any high-level causes.
2) are not orthogonal (Fig. \ref{fig:dis_cause} shows 8\% selections are ``\textbf{Multiple}'')
3) are highly imbalanced (63.6\% are change of action which can cover numerous generic event classes).
That being said, the dropdown selection in the annotation interface was actually requested by annotators - is more of an abstract reminder about what event changes roughly look like in high-level, to help eliminate common ``miss" errors.

\subsection{Motivation for Annotation Post-processing}
In Sec. \ref{sec:post-processing}, we introduced three steps to construct our Kinetics-GEBD benchmark.
Our guiding motivation for post-processing is to have automatic ways to further enhance the data quality e.g. remove very ambiguous videos (\textbf{human studies} in Table 2 tell us for consistency score $<$ 0.3, the quality is typically bad), merge very close boundaries (when two boundaries have offset $<$ 0.1s, usually there is only 1-2 frames in the between; through \textbf{human inspection} of these boundaries, we found they are typically due to annotation errors; we did not go higher thresholds like 0.2s because we did find quite some reasonable cases at 0.2s and mostly make sense for higher than 0.2s).

\subsection{Quality Assurance Mechanism}

To ensure the annotation quality: (1) Before graduating a new annotator to work on real jobs, the annotator will work on an exam batch of 100 videos and we will audit its performance.
If the performance is not satisfying, the annotator will be moved back to re-do the training curriculum.
(2) During the real annotation process, every week we conduct auditing for each annotator based on randomly sampled some of its jobs completed in that week.
If the performance is not satisfying, the annotator will be moved back to re-do the training curriculum.
If the times of re-training exceeds three, the annotator will be removed forever.
The detailed guideline of rating scores for quality assurance can be found in the Sec.~\ref{qa_guideline}.

\section{Supp: Additional Experimental Results}

\subsection{Complete Results of Precision, Recall, F1}

Table \ref{table:TAPOS_GEBD_supp} for TAPOS and Table \ref{table:Kin_GEBD_supp} for Kinetics-GEBD respectively present the detailed numbers of precision, recall, F1 score for various methods.
It is worthwhile noting that some methods like SceneDetect achieves high precision yet low recall because the method is designed to detect shot boundaries, which are easy to spot out, while misses other event boundaries.

\begin{table*}[t]
\begin{tabular}{cccllllllllll}
\multicolumn{13}{c}{(a) Precision}                                                                                                                                                                                                                                                                                                                                                                              \\ \hline
\multicolumn{2}{c|}{Rel.Dis. threshold}                                           & 0.05                      & \multicolumn{1}{c}{0.1}   & \multicolumn{1}{c}{0.15}  & \multicolumn{1}{c}{0.2}   & \multicolumn{1}{c}{0.25}  & \multicolumn{1}{c}{0.3}   & \multicolumn{1}{c}{0.35}  & \multicolumn{1}{c}{0.4}   & \multicolumn{1}{c}{0.45}  & \multicolumn{1}{c|}{0.5}            & \multicolumn{1}{c}{avg}   \\ \hline
\multicolumn{1}{c|}{\multirow{3}{*}{Unsuper.}} & \multicolumn{1}{c|}{SceneDetect} & 0.391                     & \multicolumn{1}{c}{0.506} & \multicolumn{1}{c}{0.532} & \multicolumn{1}{c}{0.576} & \multicolumn{1}{c}{0.596} & \multicolumn{1}{c}{0.608} & \multicolumn{1}{c}{0.621} & \multicolumn{1}{c}{0.628} & \multicolumn{1}{c}{0.641} & \multicolumn{1}{c|}{0.647}          & \multicolumn{1}{c}{0.575} \\
\multicolumn{1}{c|}{}                          & 

\multicolumn{1}{c|}{PA - Random}   & 0.206                     & \multicolumn{1}{c}{0.304} & \multicolumn{1}{c}{0.356} & \multicolumn{1}{c}{0.404} & \multicolumn{1}{c}{0.432} & \multicolumn{1}{c}{0.452} & \multicolumn{1}{c}{0.466} & \multicolumn{1}{c}{0.481} & \multicolumn{1}{c}{0.491} & \multicolumn{1}{c|}{0.500}          & \multicolumn{1}{c}{0.409} \\

\multicolumn{1}{c|}{}                          &
\multicolumn{1}{c|}{PA}          & 0.470                     & \multicolumn{1}{c}{0.599} & \multicolumn{1}{c}{0.662} & \multicolumn{1}{c}{0.708} & \multicolumn{1}{c}{0.740} & \multicolumn{1}{c}{0.755} & \multicolumn{1}{c}{0.771} & \multicolumn{1}{c}{0.784} & \multicolumn{1}{c}{0.795} & \multicolumn{1}{c|}{0.801}          & \multicolumn{1}{c}{0.708} \\ \hline

\multicolumn{1}{c|}{\multirow{5}{*}{Super.}}   & \multicolumn{1}{c|}{ISBA}        & 0.119                     & \multicolumn{1}{c}{0.185} & \multicolumn{1}{c}{0.230} & \multicolumn{1}{c}{0.268} & \multicolumn{1}{c}{0.301} & \multicolumn{1}{c}{0.329} & \multicolumn{1}{c}{0.356} & \multicolumn{1}{c}{0.379} & \multicolumn{1}{c}{0.392} & \multicolumn{1}{c|}{0.405}          & \multicolumn{1}{c}{0.296} \\
\multicolumn{1}{c|}{}                          & \multicolumn{1}{c|}{TCN}         & 0.140                     & \multicolumn{1}{c}{0.187} & \multicolumn{1}{c}{0.200} & \multicolumn{1}{c}{0.204} & \multicolumn{1}{c}{0.207} & \multicolumn{1}{c}{0.208} & \multicolumn{1}{c}{0.210} & \multicolumn{1}{c}{0.211} & \multicolumn{1}{c}{0.211} & \multicolumn{1}{c|}{0.211}          & \multicolumn{1}{c}{0.199} \\
\multicolumn{1}{c|}{}                          & \multicolumn{1}{c|}{CTM}         & 0.154                     & \multicolumn{1}{c}{0.197} & \multicolumn{1}{c}{0.212} & \multicolumn{1}{c}{0.221} & \multicolumn{1}{c}{0.228} & \multicolumn{1}{c}{0.233} & \multicolumn{1}{c}{0.237} & \multicolumn{1}{c}{0.242} & \multicolumn{1}{c}{0.244} & \multicolumn{1}{c|}{0.245}          & \multicolumn{1}{c}{0.221} \\
\multicolumn{1}{c|}{}                          & \multicolumn{1}{c|}{TransParser} & 0.230                     & \multicolumn{1}{c}{0.302} & \multicolumn{1}{c}{0.345} & \multicolumn{1}{c}{0.377} & \multicolumn{1}{c}{0.398} & \multicolumn{1}{c}{0.410} & \multicolumn{1}{c}{0.420} & \multicolumn{1}{c}{0.427} & \multicolumn{1}{c}{0.432} & \multicolumn{1}{c|}{0.437}          & \multicolumn{1}{c}{0.378} \\
\multicolumn{1}{c|}{}                          & \multicolumn{1}{c|}{PC}          & 0.650                     & \multicolumn{1}{c}{0.741} & \multicolumn{1}{c}{0.782} & \multicolumn{1}{c}{0.805} & \multicolumn{1}{c}{0.821} & \multicolumn{1}{c}{0.829} & \multicolumn{1}{c}{0.836} & \multicolumn{1}{c}{0.842} & \multicolumn{1}{c}{0.846} & \multicolumn{1}{c|}{0.851}          & \multicolumn{1}{c}{0.800} \\ \hline
\multicolumn{1}{l}{}                           & \multicolumn{1}{l}{}             & \multicolumn{1}{l}{}      &                           &                           &                           &                           &                           &                           &                           &                           &                                     &                           \\
\multicolumn{13}{c}{(b) Recall}                                                                                                                                                                                                                                                                                                                                                                                 \\ \hline
\multicolumn{2}{c|}{Rel.Dis. threshold}                                           & 0.05                      & \multicolumn{1}{c}{0.1}   & \multicolumn{1}{c}{0.15}  & \multicolumn{1}{c}{0.2}   & \multicolumn{1}{c}{0.25}  & \multicolumn{1}{c}{0.3}   & \multicolumn{1}{c}{0.35}  & \multicolumn{1}{c}{0.4}   & \multicolumn{1}{c}{0.45}  & \multicolumn{1}{c|}{0.5}            & \multicolumn{1}{c}{avg}   \\ \hline
\multicolumn{1}{c|}{\multirow{3}{*}{Unsuper.}} & \multicolumn{1}{c|}{SceneDetect} & \multicolumn{1}{l}{0.018} & 0.023                     & 0.025                     & 0.027                     & 0.028                     & 0.028                     & 0.029                     & 0.029                     & 0.030                     & \multicolumn{1}{l|}{0.030}          & 0.027                     \\
\multicolumn{1}{c|}{}                          
& \multicolumn{1}{c|}{PA - Random}   & \multicolumn{1}{l}{0.128} & 0.189                     & 0.221                     & 0.252                     & 0.269                     & 0.281                     & 0.290                     & 0.299                     & 0.305                     & \multicolumn{1}{l|}{0.311}          & 0.255                     \\

\multicolumn{1}{c|}{}                          & 
\multicolumn{1}{c|}{PA}          & \multicolumn{1}{l}{0.292} & 0.372                    & 0.412                     & 0.440                     & 0.460                     & 0.470                     & 0.480                    & 0.488                     & 0.494                     & \multicolumn{1}{l|}{0.498}          & 0.441                     \\ \hline

\multicolumn{1}{c|}{\multirow{5}{*}{Super.}}   & \multicolumn{1}{c|}{ISBA}        & \multicolumn{1}{l}{0.095} & 0.158                     & 0.225                     & 0.263                     & 0.296                     & 0.323                     & 0.340                     & 0.360                     & 0.373                     & \multicolumn{1}{l|}{0.386}          & 0.282                     \\
\multicolumn{1}{c|}{}                          & \multicolumn{1}{c|}{TCN}         & \multicolumn{1}{l}{0.757} & 0.940                     & 0.974                     & 0.985                     & 0.989                     & 0.990                     & 0.994                     & 0.994                     & 0.994                     & \multicolumn{1}{l|}{0.994}          & 0.961                     \\
\multicolumn{1}{c|}{}                          & \multicolumn{1}{c|}{CTM}         & \multicolumn{1}{l}{0.596} & 0.752                     & 0.811                     & 0.843                     & 0.860                     & 0.875                     & 0.886                     & 0.894                     & 0.898                     & \multicolumn{1}{l|}{0.901}          & 0.831                     \\
\multicolumn{1}{c|}{}                          & \multicolumn{1}{c|}{TransParser} & \multicolumn{1}{l}{0.386} & 0.516                     & 0.590                     & 0.642                     & 0.673                     & 0.689                     & 0.705                     & 0.714                     & 0.721                     & \multicolumn{1}{l|}{0.726}          & 0.636                     \\
\multicolumn{1}{c|}{}                          & \multicolumn{1}{c|}{PC}          & \multicolumn{1}{l}{0.436} & 0.497                     & 0.525                     & 0.541                     & 0.551                     & 0.556                     & 0.561                     & 0.565                     & 0.568                     & \multicolumn{1}{l|}{0.572}          & 0.537                     \\ \hline
\multicolumn{1}{l}{}                           & \multicolumn{1}{l}{}             & \multicolumn{1}{l}{}      &                           &                           &                           &                           &                           &                           &                           &                           &                                     &                           \\
\multicolumn{13}{c}{(c) F1}                                                                                                                                                                                                                                                                                                                                                                                     \\ \hline
\multicolumn{2}{c|}{Rel.Dis. threshold}                                           & 0.05                      & \multicolumn{1}{c}{0.1}   & \multicolumn{1}{c}{0.15}  & \multicolumn{1}{c}{0.2}   & \multicolumn{1}{c}{0.25}  & \multicolumn{1}{c}{0.3}   & \multicolumn{1}{c}{0.35}  & \multicolumn{1}{c}{0.4}   & \multicolumn{1}{c}{0.45}  & \multicolumn{1}{c|}{0.5}            & \multicolumn{1}{c}{avg}   \\ \hline
\multicolumn{1}{c|}{\multirow{3}{*}{Unsuper.}} & \multicolumn{1}{c|}{SceneDetect} & \multicolumn{1}{l}{0.035} & 0.045                     & 0.047                     & 0.051                     & 0.053                     & 0.054                     & 0.055                     & 0.056                     & 0.057                     & \multicolumn{1}{l|}{0.058}          & 0.051                     \\
\multicolumn{1}{c|}{}                          & \multicolumn{1}{c|}{PA - Random}   & 0.158          & 0.233          & 0.273          & 0.310          & 0.331          & 0.347          & 0.357          & 0.369          & 0.376          & \multicolumn{1}{c|}{0.384}          &    0.314                       \\
\multicolumn{1}{c|}{}                          & \multicolumn{1}{c|}{PA}          & \textbf{0.360} & \textbf{0.459} & \textbf{0.507} & \textbf{0.543} & \textbf{0.567} & \textbf{0.579} & \textbf{0.592} & \textbf{0.601} & \textbf{0.609} & \multicolumn{1}{c|}{\textbf{0.615}} & \textbf{0.543}             \\ \hline
\multicolumn{1}{c|}{\multirow{5}{*}{Super.}}   & \multicolumn{1}{c|}{ISBA}        & \multicolumn{1}{l}{0.106} & 0.170                     & 0.227                     & 0.265                     & 0.298                     & 0.326                     & 0.348                     & 0.369                     & 0.382                     & \multicolumn{1}{l|}{0.396}          & 0.302                     \\
\multicolumn{1}{c|}{}                          & \multicolumn{1}{c|}{TCN}         & 0.237                     & 0.312                     & 0.331                     & 0.339                     & 0.342                     & 0.344                     & 0.347                     & 0.348                     & 0.348                     & \multicolumn{1}{l|}{0.348}          & 0.330                     \\
\multicolumn{1}{c|}{}                          & \multicolumn{1}{c|}{CTM}         & 0.244                     & 0.312                     & 0.336                     & 0.351                     & 0.361                     & 0.369                     & 0.374                     & 0.381                     & 0.383                     & \multicolumn{1}{l|}{0.385}          & 0.350                     \\
\multicolumn{1}{c|}{}                          & \multicolumn{1}{c|}{TransParser} & 0.289                     & 0.381                     & 0.435                     & 0.475                     & 0.500                     & 0.514                     & 0.527                     & 0.534                     & 0.540                     & \multicolumn{1}{l|}{0.545}          & 0.474                     \\
\multicolumn{1}{c|}{}                          & \multicolumn{1}{c|}{PC}          & \textbf{0.522}            & \textbf{0.595}            & \textbf{0.628}            & \textbf{0.647}            & \textbf{0.660}            & \textbf{0.666}            & \textbf{0.672}            & \textbf{0.676}            & \textbf{0.680}            & \multicolumn{1}{l|}{\textbf{0.684}} & \textbf{0.643}            \\ \hline
\end{tabular}                                                                                                                                                                                           
\vspace{1.5em}
\caption{GEBD results on TAPOS.}
\label{table:TAPOS_GEBD_supp}
\end{table*}

\begin{table*}[t]
\begin{center}
\begin{tabular}{ccccccccccccc}
\multicolumn{13}{c}{(a) Precision}                                                                                                                                                                                                                                                                 \\ \hline
\multicolumn{2}{c|}{Rel.Dis. threshold}                                            & 0.05           & 0.1            & 0.15           & 0.2            & 0.25           & 0.3            & 0.35           & 0.4            & 0.45           & \multicolumn{1}{c|}{0.5}            & avg            \\ \hline
\multicolumn{1}{c|}{\multirow{3}{*}{Unsuper.}} & \multicolumn{1}{c|}{SceneDetect}  & 0.731          & 0.792          & 0.819          & 0.837          & 0.847          & 0.856          & 0.862          & 0.867          & 0.870          & \multicolumn{1}{c|}{0.872}          & 0.835          \\
\multicolumn{1}{c|}{}                          & \multicolumn{1}{c|}{PA - Random}    & 0.737                     & \multicolumn{1}{c}{0.884} & \multicolumn{1}{c}{0.933} & \multicolumn{1}{c}{0.956} & \multicolumn{1}{c}{0.968} & \multicolumn{1}{c}{0.975} & \multicolumn{1}{c}{0.979} & \multicolumn{1}{c}{0.981} & \multicolumn{1}{c}{0.984} & \multicolumn{1}{c|}{0.986}          & \multicolumn{1}{c}{0.938}          \\
\multicolumn{1}{c|}{}                          
& \multicolumn{1}{c|}{PA}           & 0.836          & 0.944          & 0.965          & 0.973          & 0.978          & 0.980          & 0.983          & 0.985          & 0.986          & \multicolumn{1}{c|}{0.989}          & 0.962          \\ \hline

\multicolumn{1}{c|}{\multirow{5}{*}{Super.}}   & \multicolumn{1}{c|}{BMN}          & 0.128          & 0.141          & 0.148          & 0.152          & 0.156          & 0.159          & 0.162          & 0.164          & 0.165          & \multicolumn{1}{c|}{0.167}          & 0.154          \\
\multicolumn{1}{c|}{}                          & \multicolumn{1}{c|}{BMN-StartEnd} & 0.396          & 0.479          & 0.509          & 0.525          & 0.534          & 0.540          & 0.544          & 0.547          & 0.549          & \multicolumn{1}{c|}{0.551}          & 0.517          \\
\multicolumn{1}{c|}{}                          & \multicolumn{1}{c|}{TCN-TAPOS}    & 0.518          & 0.622          & 0.665          & 0.690          & 0.706          & 0.718          & 0.727          & 0.733          & 0.738          & \multicolumn{1}{c|}{0.743}          & 0.686          \\
\multicolumn{1}{c|}{}                          & \multicolumn{1}{c|}{TCN}          & 0.461          & 0.519          & 0.538          & 0.547          & 0.553          & 0.557          & 0.559          & 0.561          & 0.563          & \multicolumn{1}{c|}{0.564}          & 0.542          \\
\multicolumn{1}{c|}{}                          & \multicolumn{1}{c|}{PC}           & 0.624          & 0.753          & 0.794          & 0.816          & 0.828          & 0.836          & 0.841          & 0.844          & 0.846          & \multicolumn{1}{c|}{0.849}          & 0.803          \\ \hline
                                               &                                   &                &                &                &                &                &                &                &                &                &                                     &                \\
\multicolumn{13}{c}{(b) Recall}                                                                                                                                                                                                                                                                    \\ \hline
\multicolumn{2}{c|}{Rel.Dis. threshold}                                            & 0.05           & 0.1            & 0.15           & 0.2            & 0.25           & 0.3            & 0.35           & 0.4            & 0.45           & \multicolumn{1}{c|}{0.5}            & avg            \\ \hline
\multicolumn{1}{c|}{\multirow{3}{*}{Unsuper.}} & \multicolumn{1}{c|}{SceneDetect}  & 0.170          & 0.185          & 0.192          & 0.197          & 0.200          & 0.202          & 0.204          & 0.206          & 0.207          & \multicolumn{1}{c|}{0.207}          & 0.197          \\
\multicolumn{1}{c|}{}                          
& \multicolumn{1}{c|}{PA - Random}    & 0.218          & 0.289          & 0.326          & 0.350          & 0.364          & 0.374          & 0.381          & 0.386          & 0.389          & \multicolumn{1}{c|}{0.393}          & 0.347          \\

\multicolumn{1}{c|}{}                          
& \multicolumn{1}{c|}{PA}           & 0.259          & 0.329         & 0.355         & 0.368          & 0.377          & 0.382         & 0.386          & 0.390          & 0.392          & \multicolumn{1}{c|}{0.395}          & 0.363          \\ \hline

\multicolumn{1}{c|}{\multirow{5}{*}{Super.}}   & \multicolumn{1}{c|}{BMN}          & 0.338          & 0.369          & 0.385          & 0.397          & 0.407          & 0.414          & 0.420          & 0.426          & 0.430          & \multicolumn{1}{c|}{0.434}          & 0.402          \\
\multicolumn{1}{c|}{}                          & \multicolumn{1}{c|}{BMN-StartEnd} & 0.648          & 0.766          & 0.817          & 0.846          & 0.864          & 0.876          & 0.885          & 0.892          & 0.897          & \multicolumn{1}{c|}{0.900}          & 0.839          \\
\multicolumn{1}{c|}{}                          & \multicolumn{1}{c|}{TCN-TAPOS}    & 0.420          & 0.508          & 0.550          & 0.576          & 0.594          & 0.609          & 0.619          & 0.627          & 0.633          & \multicolumn{1}{c|}{0.639}          & 0.577          \\
\multicolumn{1}{c|}{}                          & \multicolumn{1}{c|}{TCN}          & 0.811          & 0.894          & 0.923          & 0.938          & 0.947          & 0.952          & 0.956          & 0.959          & 0.961          & \multicolumn{1}{c|}{0.963}          & 0.930          \\
\multicolumn{1}{c|}{}                          & \multicolumn{1}{c|}{PC}           & 0.626          & 0.764          & 0.814          & 0.843          & 0.859          & 0.871          & 0.879          & 0.885          & 0.889          & \multicolumn{1}{c|}{0.892}          & 0.832          \\ \hline
                                               &                                   &                &                &                &                &                &                &                &                &                &                                     &                \\
\multicolumn{13}{c}{(c) F1}                                                                                                                                                                                                                                                                        \\ \hline
\multicolumn{2}{c|}{Rel.Dis. threshold}                                            & 0.05           & 0.1            & 0.15           & 0.2            & 0.25           & 0.3            & 0.35           & 0.4            & 0.45           & \multicolumn{1}{c|}{0.5}            & avg            \\ \hline
\multicolumn{1}{c|}{\multirow{3}{*}{Unsuper.}} & \multicolumn{1}{c|}{SceneDetect}  & 0.275 & 0.300          & 0.312          & 0.319          & 0.324          & 0.327          & 0.330         & 0.332          & 0.334          & \multicolumn{1}{c|}{0.335}          & 0.318          \\
\multicolumn{1}{c|}{}                          & \multicolumn{1}{c|}{PA - Random}    & 0.336          & 0.435          & 0.484         & 0.512          & 0.529          & 0.541          & 0.548          & 0.554          & 0.558          & \multicolumn{1}{c|}{0.561}          & 0.506         \\
\multicolumn{1}{c|}{}                          & \multicolumn{1}{c|}{PA}         & \textbf{0.396}          & \textbf{0.488} & \textbf{0.520} & \textbf{0.534} & \textbf{0.544} & \textbf{0.550} & \textbf{0.555} & \textbf{0.558} & \textbf{0.561} & \multicolumn{1}{c|}{\textbf{0.564}} & \textbf{0.527} \\ \hline
\multicolumn{1}{c|}{\multirow{5}{*}{Super.}}   & \multicolumn{1}{c|}{BMN}          & 0.186          & 0.204          & 0.213          & 0.220          & 0.226          & 0.230          & 0.233          & 0.237          & 0.239          & \multicolumn{1}{c|}{0.241}          & 0.223          \\
\multicolumn{1}{c|}{}                          & \multicolumn{1}{c|}{BMN-StartEnd} & 0.491          & 0.589          & 0.627          & 0.648          & 0.660          & 0.668          & 0.674          & 0.678          & 0.681          & \multicolumn{1}{c|}{0.683}          & 0.640          \\
\multicolumn{1}{c|}{}                          & \multicolumn{1}{c|}{TCN-TAPOS}    & 0.464          & 0.560          & 0.602          & 0.628          & 0.645          & 0.659          & 0.669          & 0.676          & 0.682          & \multicolumn{1}{c|}{0.687}          & 0.627          \\
\multicolumn{1}{c|}{}                          & \multicolumn{1}{c|}{TCN}          & 0.588          & 0.657          & 0.679          & 0.691          & 0.698          & 0.703          & 0.706          & 0.708          & 0.710          & \multicolumn{1}{c|}{0.712}          & 0.685          \\
\multicolumn{1}{c|}{}                          & \multicolumn{1}{c|}{PC}           & \textbf{0.625} & \textbf{0.758} & \textbf{0.804} & \textbf{0.829} & \textbf{0.844} & \textbf{0.853} & \textbf{0.859} & \textbf{0.864} & \textbf{0.867} & \multicolumn{1}{c|}{\textbf{0.870}} & \textbf{0.817} \\ \hline
\end{tabular}
\end{center}
\caption{GEBD results on Kinetics-GEBD.}
\label{table:Kin_GEBD_supp}
\end{table*}

\section{Supp: Detailed Quality Assurance Guideline}\label{qa_guideline}

For the simplicity of illustrating the rating instructions, we use GT in the below to stand for Ground Truth.
We refer the specific annotator as ``rater''.
As explained in the main paper, GT is not unique, depending on various human perception behaviors. 
We bear this in mind during auditing: we judge each annotated boundary via first interpreting its perception and segmentation logic and then comparing it against its corresponding correct boundary.

\subsection{Rating score definition}  

\textbf{1 = accurate} (does not have to be consistent with GT but reasonable);
\textbf{2 = minor error} (not accurate, occasional miss, the main purpose of this intermedium score level is to reflect the need of improvement while acknowledge that the rater's understanding of the annotation guideline does not have problem); 
\textbf{3 = bad, clear error} that should not happen (clear errors that have been emphasized multiple times during training, e.g. mark as shot boundary range but annotate one timestamp, mark unreasonable event boundary at the start or end of video, mark over-detailed annotations).

For the below different scenarios, an annotation for a video may fit multiple scenarios and thus result into various scores - the baddest score is deemed as the final score.

\subsection{Specific instructions for the scenario of ``Rejection''}

\textbf{1, accurate:} when the GT rejects the video while the rater marks 1 or 2 reasonable boundaries.
\textbf{2, minor:}  
when it would be desirable to annotate boundaries as GT does but the video might have some ambiguity, so that the rater rejects it;
when the GT rejects the video while the rater marks 1 unreasonable boundary.
\textbf{3, bad:} when there are clear boundaries should have been marked e.g. shot boundaries; when GT rejects the video while the rater marks more than 2 unreasonable boundaries; when the rater looks into a wrong granularity so that mistakenly considers the video contains too many boundaries or no boundaries. Recall the long jump example, marking at every step is too detailed while marking at the level of long jump is too abstract.

\subsection{Specific instructions for the scenario of ``the same boundary marked by both GT and rater''}

\head{First, check dropdown boundary cause selection i.e. shot boundary or event boundary}:
Note that in the annotation guideline, we request the rater to only select shot boundary if a boundary is both shot boundary and event boundary. If this is violated, give \textbf{3, bad} score. 
Note that we do not judge the second dropdown selection for common event boundary causes like change of subject, change of action.
For other cases, go for \textbf{1, accurate}.

\head{Second, check time difference (rater marks timestamp x seconds apart from GT video timestamp)}:
\textbf{2, minor:} x in the range of 0.3s - 0.6s. \textbf{3, bad:} x is longer than 0.6s.

\subsection{Specific instructions for the scenario of ``a clear boundary marked by GT but the rater misses it''}

Note again that if a GT boundary is reasonable to either keep or ignore, we do not consider it as a clear boundary and do not penalize the miss of it.

\textbf{2, minor:} the rater misses 25\%-50\% of all very clear boundaries (e.g. for a video of 10 clear boundaries, miss 4 very clear boundaries). \textbf{3, bad:} the video only has only a few clear boundaries while the rater misses one of them (e.g. 3 shot boundaries, miss one); the rater misses more than 50\% of all very clear boundaries (e.g. for a video of 10 clear boundaries, miss 6; for a video of 3 boundaries, miss 2).

\subsection{Specific instructions for the scenario of ``a boundary falsely marked by rater but GT would not mark it''}

Note again that if a boundary marked by rater does make sense, can be considered as reasonable 

\textbf{2, minor:} the rater marks unreasonable boundaries whose number count is 25\%-50\% of all very clear GT boundaries (e.g. for a video of 10 clear boundaries, mark 4 more additional, unreasonable boundaries).
\textbf{3, bad:} the rater marks unreasonable boundaries whose number count is more than 50\% of all very clear boundaries (e.g. for a video of 10 clear boundaries, mark 6 more unreasonable boundaries; for a video of 3 boundaries, mark 2 more unreasonable boundaries); the video only has only a few clear GT boundaries while the rater marks one more unreasonable boundaries (e.g. 3 shot boundaries, the rater marks one additional shot boundary which is not reasonable and shall not be marked).

{\small
\bibliographystyle{ieee_fullname}
\bibliography{egbib}
}

\end{document}